%% file: main.tex
\documentclass[a4paper,plainchapterheads,yschapters,truedoublelespace,openright]{iitbthesis}
\usepackage{epsfig}
\usepackage[T1]{fontenc}
\usepackage[utf8]{inputenc}
\usepackage{mathptmx}
\usepackage{amssymb}
\usepackage{amsmath,epsfig}
\usepackage{graphicx,graphics}
\usepackage{url}
\usepackage{hyperref}
\usepackage[capitalise]{cleveref}
\usepackage{textcomp}
\usepackage{enumitem}
\usepackage[square,numbers]{natbib}
\usepackage{pdflscape}
\renewcommand\bibname{References}
\usepackage{titlesec}
\usepackage{bm,bbm}
\usepackage{float,lscape}
\usepackage{subfig}

\usepackage{epstopdf}
\usepackage{tabu}
\usepackage{multirow}
\usepackage{array}
\usepackage{booktabs}
\usepackage{graphicx}
\usepackage{caption}
\usepackage{lettrine}
\usepackage{enumitem}
\usepackage{lscape}
\usepackage{rotating}
\usepackage{booktabs}
\usepackage{listings}
\usepackage{longtable}
\usepackage{anyfontsize}
\usepackage{t1enc}
\usepackage{relsize}
\usepackage{placeins}
\usepackage{nomencl}
\usepackage{rotating}
\usepackage{glossaries}
\usepackage{scrlayer}
\usepackage{enumitem}
\setlist{nolistsep,leftmargin=*}
\usepackage{xcolor}
\usepackage{courier}
\usepackage[linesnumbered,ruled,vlined]{algorithm2e}

\usepackage[titletoc]{appendix}

\DontPrintSemicolon

\SetKwComment{Comment}{\color{green!50!black}// }{}

\SetKwProg{Function}{function}{}{}

\begin{document}

\abovedisplayskip=1mm
\abovedisplayshortskip=1mm
\belowdisplayskip=1mm
\belowdisplayshortskip=1mm

\include{prelude}

\setlength{\parskip}{1.5mm}
\setlength{\parindent}{4mm}
\renewcommand{\baselinestretch}{1.0}

\titlespacing{\chapter}{0cm}{0mm}{0mm}
\titleformat{\chapter}[display]
  {\normalfont\huge\bfseries\centering}
  {\chaptertitlename\ \thechapter}{1pt}{\Huge}
  \titlespacing*{\section}
  {0pt}{4mm}{4mm}
  \titlespacing*{\subsection}
  {0pt}{4mm}{4mm}
\pagenumbering{arabic}

\makeatletter
\def\cleardoublepage{\clearpage\if@twoside \ifodd\c@page\else
	\hbox{}
	\vspace*{\fill}
	\begin{center}
		This page was intentionally left blank.
	\end{center}
	\vspace{\fill}
	\thispagestyle{empty}
	\newpage
	\if@twocolumn\hbox{}\newpage\fi\fi\fi}
\makeatother

\include{chap_01_Introduction}
\include{chap_02_Review_of_literature}
\include{chap_03_Method_Details}
\include{chap_05_expts_results_and_analysis}
\include{chap_06_efficiency}
\include{chap_07_Summary_and_Conclusions}


\begin{appendices}
	\include{Appendix_I}

	\include{Appendix_II}

\end{appendices}

\setlength{\parskip}{5mm}
\titlespacing{\chapter}{0cm}{0mm}{0mm}
\titleformat{\chapter}[display]
  {\normalfont\huge\bfseries}
  {\chaptertitlename\ \thechapter}{1pt}{\Huge}

\bibliographystyle{References/elsarticle-harv}
\bibliography{References/mybibfile}



\newpage
\setlength{\parskip}{2mm}
\titlespacing{\chapter}{0cm}{0mm}{0mm}
\titleformat{\chapter}[display]
  {\normalfont\huge\bfseries \centering}
  {\chaptertitlename\ \thechapter}{20pt}{\Huge}

\chapter*{Acknowledgments}
\label{ch:Acknowledgments}
\addcontentsline{toc}{chapter}{\nameref{ch:Acknowledgments}}
\vspace{10mm}
I wish to record a deep sense of gratitude to \textbf{Prof. Preethi Jyothi}, my supervisor, for her valuable guidance and constant support in all stages of my research. I would also like to thank everyone at STCI at Microsoft IDC, especially Basil and Niranjan, for their support, resources, and time over the 15 months of work put in this project and the previous one.

Since this is also my last project from IIT Bombay, I would like to thank my friends for making life at college interesting and helping me focus on what's important.

My most sincere gratitude goes to my parents, for tolerating me over the phone during my time at college and at home during the pandemic.

\end{document}

%% file: prelude.tex


\pagenumbering{gobble}

\title{Gujarati-English Code-Switching Speech Recognition using ensemble prediction of spoken language}
\author{Yash Sharma}
\date{May 21, 2021}

\rollnum{17D070059} 

\iitbdegree{Bachelor of Technology}

\project

\department{Computer Science and Engineering}

\setguide{Prof. Preethi Jyothi}
\setexguide{Basil Abraham, Microsoft IDC Hyderabad}

\maketitle

\clearpage
\pagenumbering{roman}
\begin{abstract}
  \input{abstract}

\end{abstract}

\tableofcontents



%
%


%% file: abstract.tex
\renewcommand{\thepage}{\roman{page}} \setcounter{page}{1}

An important and difficult task in code-switched speech recognition is to recognize the language, as lots of words in two languages can sound similar, especially in some accents. We focus on improving performance of end-to-end Automatic Speech Recognition models by conditioning transformer layers on language ID of words and character in the output in an per layer supervised manner. To this end, we propose two methods of introducing language specific parameters and explainability in the multi-head attention mechanism, and implement a Temporal Loss that helps maintain continuity in input alignment. Despite being unable to reduce WER significantly, our method shows promise in predicting the correct language from just spoken data. We introduce regularization in the language prediction by dropping LID in the sequence, which helps align long repeated output sequences.

%% file: chap_01_Introduction.tex
\chapter{\textsc{Introduction}}
\section{Background and Previous Work}

\label{sec:background}

Our task was to improve Automatic Speech Recognition performance on code-switched Gujarati-English data, using 100 hours of monolingual Gujarati and 100 hours of code-switched Gujarati-English with Gujarati as the matrix language. This training data is our baseline, as we use other variations of added data as intuition suggests. As a baseline and a starting point for further experiments, we use an end-to-end ASR model: a Transformer based encoder-decoder hybrid network \cite{watanabe2017hybrid} with both label (smoothing) loss and CTC loss\cite{graves2006connectionist}; with a beam search based decoding on the CTC output, decoder output and a RNN based character language model. More details are in Chapter \ref{ch:expts}.

In order to improve the WER of the model, we look at transliteration first. We built a WFST-based \cite{karimi2011machine,knight1997machine} English to Gujarati (and hence the reverse) transliteration model based on a common phoneme set and a large lexicon in both languages.

The monolingual Gujarati data is pruned through the inverse WFST, transliterating words to English for which a parallel word exists. We then ran a number of experiments with different amounts of transliterated data and tabulated our results, a summary of which is in table \ref{tab:trans_results}.

\setlength{\tabcolsep}{15pt}
\renewcommand{\arraystretch}{0.5}
\begin{table*}[t!]
    \caption{WER(CER) in \%: Transliteration Experiments}
    \centering
    \begin{tabular}{l r r r}
    \toprule
    \textbf{System} & \textbf{test-cs} & \textbf{dev-mono} \\
    \midrule
    \textit{Baseline} \\
    \midrule
    Guj 100 + CS 100 & \textbf{32.5} (\textbf{16.4}) & \textbf{38.0} (\textbf{21.1}) \\
    \midrule
    \textit{Sequential - retraining}\\
    \midrule
    mono, CS & 34.7 (17.5) & 45.4 (28.0) \\
    mono, trans, CS & 34.4 (17.3) & 46.1 (28.5)\\
    trans, CS & 34.6 (17.4)& 45.9 (28.9)\\
    \midrule
    \textit{Sequential - pretraining} \\
    \midrule
    trans, CS & 34.7 (17.6) & 44.7 (27.6)\\
    mono, trans & 52.9 (38.9)&41.8 (22.6)\\
    \midrule
    \textit{Mixed data} \\
    \midrule
    mono + exp-trans & 51.2 (38.4) & 41.9 (23.1)\\
    mono + exp-trans + CS & 33.0 (21.5) & 38.6 (16.8)\\
    \midrule
    \textit{Transliteration Modes} \\
    \midrule
    trans + CS &	33.4 (16.7) & 45.5 (27.2)\\
    ran-trans + CS & 33.7 (17.1) & 45.8 (27.7)\\
    exp-trans + CS & 33.5 (16.9) & 45.8 (25.1)\\
    \midrule
    \textit{Using CS + Guj LM}\\
    \midrule
    trans, CS & 34.0 (17.2) & 43.0 (26.0)\\
    exp-trans + CS & 33.1 (16.9) & 42.7 (25.1)\\
    \bottomrule
    \end{tabular}
    \label{tab:trans_results}
\end{table*}

\section{Motivation}

We suspected that the baseline model is actually better at phoneme recognition than the word error rate showed. We hypothesize that the model is predicting some words in one language (e.g. proper nouns), while the reference for that word is in the other language. We measured the WER after transliterating both reference and prediction text to a common language using a standard transliteration library \cite{Bhat:2014:ISS:2824864.2824872}, the WER reduced significantly. We even used a phoneme based transliteration (to the same phoneme set used for WFST construction), which gave us the same results. This confirms our hypothesis. We need to improve the model's prediction of the correct language, while keeping it's original capacity to translate speech to phonemes. The results on a test set is in table \ref{tab:normwer_prev}.

\begin{table*}[t!]
    \caption{Normalized WER in \%}
    \centering
    \begin{tabular}{l r r}
    \toprule
    \textbf{System} & norm. mode & \textbf{test-cs} \\
    \midrule
Guj 100 + CS 100 & regular & 32.5\\
 & phone norm. & 31.6 \\
 & guj WER & 32.0 \\
 & eng WER & \textbf{31.5} \\
 \midrule
exp-trans + CS & regular & 33.2\\
 & phone norm. & 32.2 \\
 & guj WER & 32.7 \\
& eng WER & \textbf{32.1} \\
 \midrule
Guj 100 & regular & 51.4 \\
 & phone norm. & 49.4\\
 & guj WER & 48.5 \\
 & eng WER & \textbf{47.8} \\
 \bottomrule
 \end{tabular}
\label{tab:normwer_prev}
\end{table*}

\section{Objective \& Method Summary}
Our objective was to improve WER by better understanding the true language of the spoken words, and try to achieve post-transliterated WER level performance.
We attempt to improve language identification and therefore prediction by conditioning the encoder (and the decoder) to a predicted language ID for each frame in the speech (and text) encoding. We train the parameters with an additional temporal classification technique mapping from language identity probability distribution over the temporal encodings to language IDs of each word or character in the output.

%% file: chap_02_Review_of_literature.tex
\chapter{\textsc{Related Work}}
\label{sec:2}
We'll discuss two papers, one in the Multilingual Neural Machine Translation task, and the other targeting code-switched speech recognition using a bi-encoder model. There are many papers in this domain tackling a similar issues \cite{dabre2020survey,sitaram2019survey}, but we'll focus on just these two specifically, as they resonate the closest to our contribution that follows.

\section{Multilingual Neural Machine Translation}

\begin{figure}
    \centering
    \includegraphics[scale=0.3]{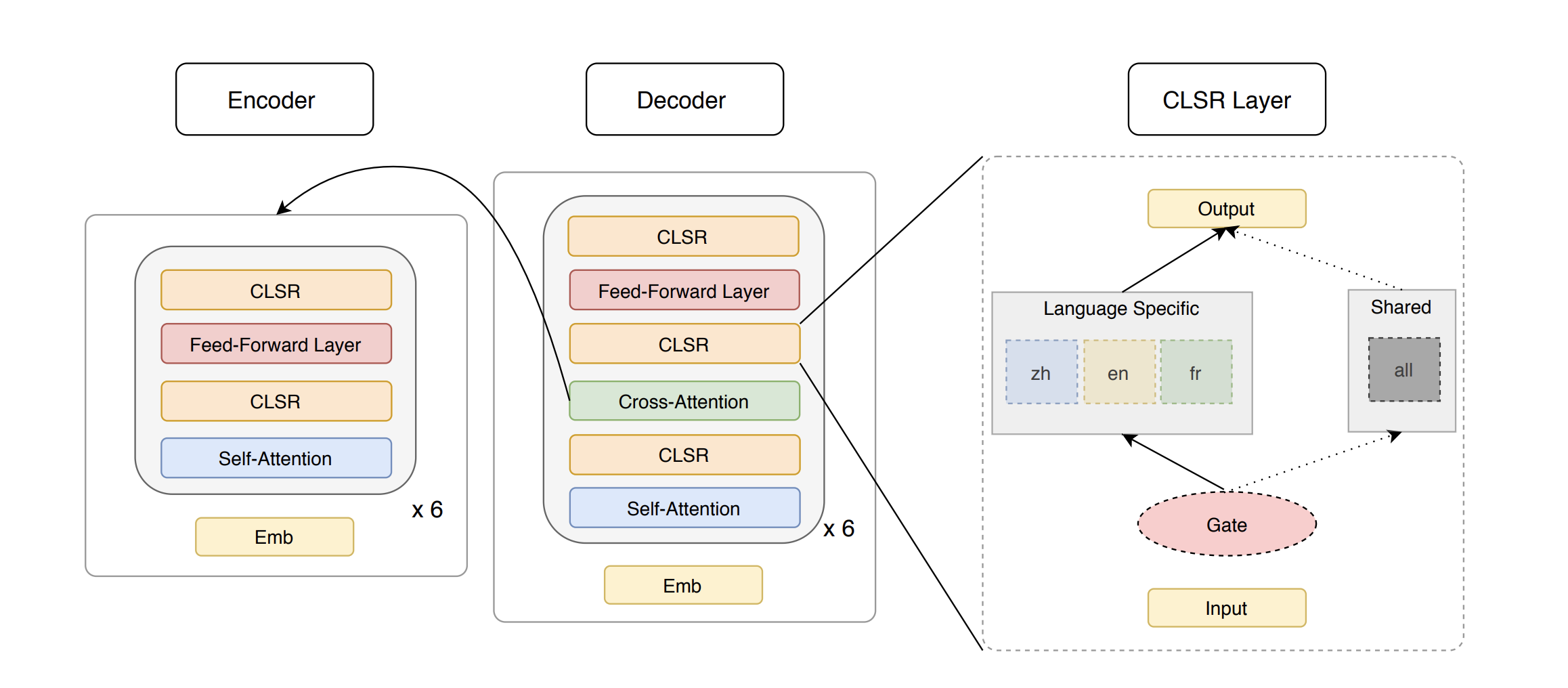}
    \caption{Language conditioned Neural Machine Translation with parameter sharing}
    \label{fig:clsr}
\end{figure}

Zhang et al. \cite{zhang2021share} in their work use language specific and agnostic characters as an alteration to a machine translation model, which makes use of an additional language ID tokens augmented to their text in order to make use of language specific and language agnostic parameters (see figure \ref{fig:clsr})

\begin{align*}
    z^{l+1} &= \text{LN}(z^l + \text{CLSR}(f(z^l))) \\
    \text{CLSR}(f(z^l)) &= g(z^l) \mathbf{h}^{\text{lang}} +  (1 - g(z^l))\mathbf{h}^{\text{shared}} \\
    \text{with}\ \ \mathbf{h}^\text{lang} = &f(z^l) W^\text{lang},\ \ h^\text{shared} = f(z^l) \\
    g(z^l) = \sigma(G(z^l) + \alpha(t)&\mathcal{N}(0, 1)),\ G(z^l) = \text{Relu}(z^l \mathbf{w_1} + \mathbf{b}) \mathbf{w_2}
\end{align*}

A similar method can be used this for separate language specific parameters based on input speech. A common observation is that even if someone does not understand the language another person is speaking, if they've heard it before, they'll be able to identify the language. One could even pick out words that come from a common language that both people know.

\section{Bi-Encoder Code-switched Speech Recognition}

Lu et al. \cite{lu2020bi} attempt to improve performance on English-Mandarin code-switched speech recognition using a dual encoder model. They first pretrain each encoder on separate models with separate data - monolingual data coming from each language. Then, treating these encoders as ``experts", they use an interpolation of the respective encodings using a weight also calculated frame by frame using the same encoding, with additional trainable parameters. The new parameters for this weight are tuned by direct optimization on code-switched data. They call this a mixture of experts model.

This leads us to a new idea - add language-ID context, especially on parameters of the model which have only seen the acoustic features of the input. 
We derive ideas from papers blah blah citations here, and noticed that a lot of papers seem to model language dependence by making separate decoders, or separate set of models and then get a weighted average of the outputs, like experts in different languages accumulating their opinion on something.

So the language conditional model has to keep a few things in mind. The model's number of parameters should not scale up too much, and should utmost be proportional to a constant times the number of languages in question. This proportionality constant should be much less, 1-5\% of the number of parameters in the baseline, in order to be comparable in capacity. For e.g. if we just decide to add two encoders to the model for two languages, this would blow up the parameters by close to 50\%, in which case we'll also have to compare this new model with performance from a bigger model, which is what Lu et al. \cite{lu2020bi} do in their work.

%% file: chap_03_Method_Details.tex
\chapter{\textsc{Method Details}}
\label{sec:3}
We use the example of a self-attention network for these details \cite{vaswani2017attention}. For each attention layer - we already have 8 multi-head attention units - each having roughly 3 million parameters of the form query, key and value embeddings. We propose that in "gating" layers, a modified version we propose, we have separate q,k,v embeddings for each language, in each head. The encoding from the previous layer is passed to each attention head's each language segment, and fed through the FF embeddings.

We do a linear interpolation of our language specific embeddings, where the weights for the interpolation comes from a special feedforward gating layer, which is fed the same encoding from the previous layer.

\section{Regular Self Attention}
\label{sec:regular_attention}

Let $z^{l-1}$ be time series encoding from the previous layer. Let $h$ be the number of layers

\begin{align*}
    \forall i \in [h]\ \text{ head}_i &= \text{Attention}(z^{l-1} Q^l_i, z^{l-1} K^l_i, z^{l-1} W^l_i) \\
    z^l = \text{Con}&\text{cat}(\text{head}_1, \text{head}_2, ..., \text{head}_h) W^l
\end{align*}

We ignore layer normalization and residual connections, as they are not relevant for this discussion.

\section{Method 1 - Pre-Attention Addition}
Let $L$ be the number of languages. We do the linear interpolation before the multiplication and softmax. Once the interpolated query, key and value are obtained, the regular process of attention calculation is continued.

\begin{align*}
    g^l_1, g^l_2, ..., g^l_L &\xleftarrow[]{} G^l(z^{l-1}) \\
    \forall 1 \leq t \leq T &\sum\limits_{n=1}^L g^l_{nt} = 1 \\
    \forall n \in [L]\ \ q^l_{in} = z^{l-1} Q^l_{in},\ 
                       k^l_{in} &= z^{l-1} K^l_{in},\ 
                       v^l_{in} = z^{l-1} V^l_{in}\\
                        q^l_i = \sum\limits_{n=1}^L g^l_n * q^l_{in},\ 
                        k^l_i &= \sum\limits_{n=1}^L g^l_n * k^l_{in},\ 
                        v^l_i = \sum\limits_{n=1}^L g^l_n * v^l_{in} \\
    \forall i \in [h]\ \text{ head}_i &= \text{Attention}(q^l_i, k^l_i, v^l_i) \\
    z^l = \text{Concat}(\text{he}&\text{ad}_1, \text{head}_2, ..., \text{head}_h) W^l
\end{align*}

\section{Method 2 - Post-Attention Addition}
\label{sec:gating_method2}

A modification of this was tested. Instead of interpolating all Q, K, V individually, we first let the individual language attend vectors and calculate new encodings, and then interpolate these encodings.

\begin{align*}
    \forall n \in [L]\ \ q^l_{in} = z^{l-1} Q^l_{in},\ 
                      k^l_{in} &= z^{l-1} K^l_{in},\ 
                      v^l_{in} = z^{l-1} V^l_{in}\\
    \forall i \in [h],\ \forall n \in [L]\ \text{ head}_{in} &= \text{Attention}(q^l_{in}, k^l_{in}, v^l_{in}) \\
    \forall n \in [L]\ \text{Lhead}_n &= \text{Concat}(\text{head}_{1n}, \text{head}_{2n}, ..., \text{head}_{hn}) \\
    \forall n \in [L]\ g^{'}_n &= G^l(\text{Lhead}_n) \\
    \forall n \in [L]\ g_n &= \text{Softmax}_n(g') \\
    \forall i \in [h]\ \text{ head}_i &= \sum\limits_{n=1}^L g_n * \text{head}_{in} \\
    z^l = \text{Concat}(\text{head}_1&, \text{head}_2, ..., \text{head}_h) W^l
\end{align*}

The difference in cross-attention will come only in method 1, where the gating weights will be calculated separately for key and value using the encoder encoding, and for query using the output embedding. The same set of feed-forward parameters are used for all.

These gating weights are then trained jointly with the ASR network by using word-level or character level language ID of each example. A reference language ID sentence is passed, which would have the same length as of the output text if it's character-level (including spaces) or would have the length of the number of words in the reference. In the case of word-level, we're assuming that mixing of languages within words doesn't occur. How this loss is designed is covered in Section \ref{gating_loss}.

We run a number of experiments with these methods of training gating parameters, both by jointly training parameters in $G^l$ vs disconnecting it from the end-to-end network and training it separately (treating the weights themselves as constants during back-propagation, and updating them separately).

We also experimented with unified parameters for each encoder layer (i.e. $\forall l\ G^l = G$), as well as separate gating for every layer, and we discarded the former because \textbf{a)} we wanted to give freedom for each layer to attended for different parts of encoding differently, \textbf{b)} encoding passed to each layer is different anyway, so same gating parameters for all layers may not fit well, and \textbf{c)} each gating layer $G^l$ is only a few thousand parameters (for a hidden dimension size 1024, it's $1024 * L$ number of parameters), so this does not significantly change the number of parameters in the model.

The number of parameters are majorly changed due to having separate feedforward parameters $Q^l_{in}$ etc. instead of $Q^l_i$. We notice a 1.1\% increase in the total number of parameters in the ASR model per layer. This is significantly once multiplied with 18 layers in encoder + decoder, so in later experiments we add gating only to a few layers of the encoder and decoder.

All experimental results are covered in Section \ref{sec:results}

%% file: chap_05_expts_results_and_analysis.tex
\chapter{\textsc{Training, Results, and Analysis}
\label{ch:expts}}
\noindent In order to train the Language ID identification in the model in a supervised fashion, we come up with a series of methods that we will cover in this next section.

We then present a number of experiments utilizing these techniques, and the Character Error Rate (CER) and Word Error Rate (WER) of every experiment on a 2 hour code-switched test set.

\section{Gating Loss}
\label{gating_loss}

Each spoken speech encoded frame can be associated with a language (or a probability distribution over which language it belongs to) when we listen to a code-switched speech. The gating weights that interpolate the attention encodings represent exactly this. Additional tokens like $\langle\text{SPACE}\rangle$, $\langle\text{UNK}\rangle$, $\langle\text{BLANK}\rangle$, $\langle\text{EOS/SOS}\rangle$ are ignored. This is reasonable since we're not actually looking to match the output sequence exactly, spaces and everything, but just maximizes the likelihood of this output sequence and associate the language ID series with continuous parts of the speech. However due to difference in temporal resolution, this cannot be directly aligned with the language IDs, either character based or word-based.
In addition to this every encoder (and decoder) layer produces its own set of probability distribution (henceforth referred to as gating weights) that needs to be aligned, and not just one set of gating weights for each example passed through the model.

We implement a temporal alignment of each of the layer's gating weights independently. Let's take an example of a 20 length input sequence, and a 4 length output sequence. For the output sequence $\langle \text{GUJ}\rangle$ $\langle\text{GUJ}\rangle$ $\langle\text{ENG}\rangle$ $\langle\text{GUJ}\rangle$, one encoder layer can have most likely weights \linebreak $\langle\text{GUJ}\rangle^{10}\ \langle\text{ENG}\rangle^{3}\ \langle\text{GUJ}\rangle^{7}$, and another layer can have $\langle\text{GUJ}\rangle^{9}\ \langle\text{ENG}\rangle^7\ \langle\text{GUJ}\rangle^{4}$. An unacceptable alignment would be $\langle\text{GUJ}\rangle^{1}\ \langle\text{ENG}\rangle^2\ \langle\text{GUJ}\rangle^{13} \langle\text{ENG}\rangle^{4}$, as it does not have sufficient attention to Gujarati in the beginning of the encoding, and has attention to English parameters at the end. Each layer has independence of how much of which part of speech it wants to attend to, while also maximizing alignment likelihood with the output sequence. Hence in each method of alignment, the loss is independently calculated for each layer, and summed/averaged over the gating layers.

We use two representations of language ID per utterance - word level and character level. Word level sequences are short and don't have $\langle\text{SPACE}\rangle$, only one language token per word. Character-level sequences are as long as the character sequence output themselves, with $\langle\text{SPACE}\rangle$ tokens as delimiters between character LID tokens.

\subsection{CTC Loss}
\subsubsection{With Projection}
\label{sec:ctc_wproj}

This is the simplest of the examples. When aligning any multi-feature distribution to a probability distribution over a vocabulary based time series, \textbf{Connectionist Temporal Classification} (CTC) is the most commonly used loss and alignment mechanism\cite{graves2006connectionist}.

$$ CTCLoss(\mathbf{x}, \mathbf{y}) = -\text{log}(P(\mathbf{y} | \mathbf{x})) = -\text{log}\big( \sum\limits_{\mathbf{a} \in B^{-1}(\mathbf{y})} P(\mathbf{a} | \mathbf{x})\big)$$

$B^{-1}(y)$ is the inverse of the two-step projection: (a) remove all consecutive repeating characters - (b) remove $\epsilon$

The features (of time series length N, and features D) are first projected to a vocabulary dimension V with trainable feedforward parameters, and taken softmax over each time frame independently (this is the independece assumption). The obtained probability distribution is vector \textbf{a}, which is then aligned to output sequence \textbf{y} using the CTC loss function. An advantage of this is we can have a probability distribution over all the output tokens (non-language ones too, as discussed in the next method), projecting from just two features - gating weights for each language. A disadvantage is that the gating weights may not directly correspond to the language IDs in this case, depending on how the parameters end up getting trained.

\subsubsection{Without Projection}

But why use projection when we already have a probability distribution from out features \textbf{x} (gating weights) over a subset of the vocabulary? Our vocabulary consists of the language IDs for each language in the task, a space token $\langle\text{SPACE}\rangle$, a blank token $\langle\text{BLANK}\rangle$, unknown token $\langle\text{UNK}\rangle$, and a common token for end and start of sentence $\langle\text{EOS/SOS}\rangle$. 

So for deriving \textbf{a} from \textbf{x}, we make a smoothing parameter $\alpha$, give probability \\\texttt{a[i] $= \alpha$ * x[i]} $\ \forall i \in \text{language tokens}$, and \texttt{a[i']} $= (1 - \alpha) / 4\ \ \forall \texttt{i'} \in \text{remaining tokens}$.\\ This means that the most likely output sequence will only have tokens from one of the languages (for $\alpha > 0.33$). We just want a close approximation of which parts of speech should be encoded by which expert - in this case, a choice between Gujarati and English.

\subsection{Label Smoothing Loss (for the decoder, char-based LID only)}
\label{sec:label_loss}
For the cross attention in the decoder, the encoding fed to the query FF is the output embeddings, whereas the key and value is the speech encoding from the last encoder layer. While the latter two have the length of the processed speech, the output embedding has the length of the output text (number of characters). Hence the gating weights also have the same resolution as a character-based LID sequence. Therefore, cross-entropy per time frame can be used, and we use KL Divergence of the two probability distribution as a proxy for cross-entropy, with a small smoothing factor.

Again there are two methods to go about this, with and without projection, from these gating weights to the vocabulary discussed in the previous method.

\subsection{Seamless Temporal Classification (STC) Loss}
\label{sec:stc_loss}

As mentioned in the sub-section \ref{sec:label_loss}, the ideal one-to-one cross-entropy loss is a very particular case, it can't be used for the rest of the network.
In addition to this, there are some issues with the current way CTC is implemented, especially in the context we're using it.

\begin{enumerate}
    \item In order to align with the output sequence with repeated tokens, the input sequence needs to predict blank tokens $\epsilon$, (represented earlier as $\langle\text{BLANK}\rangle$) in between. For e.g. if output is ``aaaaabbbabbab", then one of the input alignments is ``a$\epsilon$a$\epsilon$a$\epsilon$a$\epsilon$ab$\epsilon$b$\epsilon$bab$\epsilon$bab", and this input has the minimum number of $\epsilon$s that are needed. This is not ideal for predicting languages per speech frame, as such frequent interruptions between characters do not make sense.
    \item Due to the requirement of blank tokens, an output of length $N$, with a lot of repetitions will require up to $2N-1$ tokens in the input that needs to align with it. This is especially true for the 50\% of our data, which is fully monolingual, and the output will be all <GUJ> (with spaces in between). This issue can be overstepped in the case of word-based LID mapping, but is considerable in char-based LID mapping. This requirement is not necessarily met by the speech encodings in the encoder, and definitely not by the output embeddings in the decoder (which have the same length $N$. In this case, when there are no inputs possible to align with the output, the likelihood summation of CTC is 0, and hence the loss is positive infinite.
\end{enumerate}

The reason why CTC requires $\epsilon$ is so that the projection function over all input alignments $\in (\Sigma\cup \{\epsilon\})^*$ is an onto function on $\Sigma ^*$. If there were no blank tokens, the model will fail to predict consecutive repeated tokens in it's output space.

So we propose to get rid of the blank tokens for the meantime. For an output sequence, the model should aim at predicting input that aligns with the output by stretching the latter with different intensities at different points of it's temporal space. We maximize the likelihood of all such input alignments, each log likelihood normalized with all possible other outputs they can align to.

\newcommand{\B}{\mathbb{B}}
Formally, 
$$\B^{-1}_n(\mathbf{y} = a_1a_2...a_m) = \{\mathbf{a} = (a_1)^+ (a_2)^+...(a_m)^+ \big\rvert |\mathbf{a}| = n \}$$ is the inverse of projection in STC formulation. Treat $(x)^+$ as a regular expression, i.e., x repeated at least once. We fix the length of the input as $n$ for a specific inverse function, as it goes unsaid in the likelihood formula for CTC loss as well.

As a difference to CTC, we do not treat the inverse of $\B^{-1}$ as a function because we no longer can. Instead it is a set of outputs stemming from a single high temporal input $\mathbf{a}$.
$$\B(\mathbf{a} = (a_1)^{n_1}(a_2)^{n_2}...(a_k)^{n_k}) = \{ \mathbf{y} = (a_1)^{m_1}(a_2)^{m_2}...(a_k)^{m_k} \big\rvert \forall i\ m_i \leq n_i,\ m_i \geq 1\}$$
where all consecutive $a_i$ are distinct, and $n_i \geq 1$ is how many times $a_i$ is repeated.

Thus we have $|\B(\mathbf{a})| = \Pi_{i=1}^k n_i$, the small output is of length $k$, and the largest is of length $\Sigma_{i=1}^k n_i$

The STC loss for input x and output y hence is
$$ STCLoss(\mathbf{x}, \mathbf{y}) = -\text{log}(P(\mathbf{y} | \mathbf{x})) = -\text{log}\big( \sum\limits_{\mathbf{a} \in \B_n^{-1}(\mathbf{y})} \frac{P(\mathbf{a} | \mathbf{x}) }{\mathbf{|}\B(\mathbf{a})\mathbf{|}} \big)$$

where $n = |\mathbf{x}|$, $P(\mathbf{a} | \mathbf{x})$ is the same as before, with the independence assumption.
The normalization $|\B(\mathbf{a})|$ allows us to define the quantity inside the log as a probability P(y | x).

The removal of blank tokens results in the inability to form an onto mapping to $\Sigma^*$, as we no longer have a single "projection" from the most likely input sequence. From the current formulation, each of the output sequences from $\B(\mathbf{a})$ are deemed equally likely.

However, in our task, we just want each encoder to attend to language specific parameters for specific parts of the encodings, and we want each layer to have it's own freedom in how long it wants to attend to which part of the sequence. We also do not wish to check the most likely language characters, as this is reflected also from the actual character output of the complete end-to-end model. We do however, do a qualitative analysis of the most likely language IDs from each layer, and also during the beam search via the decoder, in the temporal resolution of the speech, which we will discuss in the later sections.

STC Loss can be presented again with or without projection, but we focused on experiments only without projection. This loss is then averaged from each layer, and added to the hybrid loss using a weight ($w_g = 1.0$ if unspecified)

\subsection{CTC Loss with trimming}

We discussed a flaw of the CTC loss while introducing STC loss - in output sequence such as in our task with large amounts of character (LID) repetition, we may require up to twice the number of time frames in the input sequence, which is not practically enforced by the model when pre-processing the acoustic frames with 2D convolutions and mel spectrogram calculation.

However, if there \textit{is} a lot of repetition in the output sequence, that also brings redundancy. If we could trim down these sequences randomly dropping some frames enough so the input can fit the modified output, they would still roughly maintain their semantic form. Hence a simple algorithm is formulated to conservatively trim the output sequence until its trivial (single frame output) or fits in the input length. The latter is kept constant throughout.

The random trimming of output sequences on the fly further acts as a regularizer over several epochs, which helped validation loss in some experiments with this method go down upto twice as many epochs as regular baseline or gating methods. The disadvantage is this increases training time per epoch, due to the conservative trimming and retries as is evident from the code.

\section{Experiments and Results}
\label{sec:results}
Despite all the modifications proposed in our work, we were unable to beat the performance of the baseline model. As defined in the previous semester, 100 hours of monolingual Gujarati speech 100 hours of code-switched English-Gujarati speech (with text) put together was the ideal as well as realistic mixture of supervised data for our experiments. Additionally, the LM (used for beam-search decoding) is also trained on the same text). In experiments where we add additional data or change the dataset for ASR training, we keep the LM same as before.

The training is annotated with 12 hours of validation dataset comprising of code-switched data. The model is evaluated on 3 different datasets - a 2 hour subset of this validation set referred to as \textbf{d-cs}, an unseen 2 hour code-switched test set (\textbf{t-cs}) and an unseen 2 hour monolingual dataset (\textbf{d-mono}).

Training takes around 40 epochs to complete, however some experiments involving trimming based regularization took 60 or more epochs to converge.

First to make sure the baseline results are reproducible despite the changes in code over the last few months, we redo the baseline experiment. The architecture is as discussed before (Section \ref{sec:background}, \cite{watanabe2017hybrid}). There are 12 encoder layers and 6 decoder layers, each with 1024 hidden units, and 8 attention heads. We then start  two experiments, one with joint gating parameters ($G^l = G$) and the other separate. They use CTC Loss with projection (ref: section \ref{sec:ctc_wproj}). The language ID is word-based. The projection parameters are shared. The gating parameters in both these experiments are disconnected from the network and trained separately (see table \ref{tab:results1}).

\setlength{\tabcolsep}{5pt}
\begin{table*}[t!]
    \caption{WER in \%: Method 1 - Share or not?}
    \centering
    \begin{tabular}{c l r r r}
    \toprule
    \textbf{S.no}	&	\textbf{Name}	&	\textbf{d-cs}	&	\textbf{t-cs}	&	\textbf{d-mono}	\\
    \midrule
    1	&	(new) Baseline	&	\textbf{32.6}	&	32.9	&	\textbf{38.3}	\\
    2	&	Shared gating	&	33.6	&	33.4	&	39	\\
    3	&	Multi gating	&	32.8	&	\textbf{32.8}	&	38.4	\\
    \bottomrule
    \end{tabular}
    \label{tab:results1}
\end{table*}

Sharing parameters was not justified (section \ref{sec:gating_method2}), so all further experiments shed this condition. Also there were very few projection parameters (in general, it would be $\text{num\_lang} * (\text{num\_lang} + 4) $, we also separate these parameters for each layer. We also try \textbf{joint training} of the gating parameters with the rest of the E2E ASR model, i.e., no disconnection (see table \ref{tab:results2}).

\begin{table*}[t!]
    \caption{WER in \%: Method 1 - multi projection and joint training}
    \centering
    \begin{tabular}{c l r r r}
    \toprule
    \textbf{S.no}	&	\textbf{Name}	&	\textbf{d-cs}	&	\textbf{t-cs}	&	\textbf{d-mono}	\\
    \midrule
    1	&	Baseline	&	32.6	&	32.9	&	38.3	\\
    4	& Gating with multi-proj CTC	&	33.7	&	33.6	&	39.0	\\
    5	& multi-proj CTC and joint training	&	33.4	&	33.3	&	39.2	\\
    \bottomrule
    \end{tabular}
    \label{tab:results2}
\end{table*}

A missing piece in this that the parameters attend to the probability distribution per time series only, and independently. We try another experiment where the projection takes a fixed context ($k = 5$ for e.g.) of gating weights ($t - k$ to $t + k$). Further to this, we add a smoothing loss to the experiment, as we see that the language will not change for several frames at a time (see table \ref{tab:results3}

\begin{table*}[t!]
    \caption{WER in \%: Method 1 - Fixed context vs smoothing}
    \centering
    \begin{tabular}{c l r r r}
    \toprule
    \textbf{S.no}	&	\textbf{Name}	&	\textbf{d-cs}	&	\textbf{t-cs}	&	\textbf{d-mono}	\\
    \midrule
    1	&	Baseline	&	\textbf{32.6}	&	\textbf{32.9}	&	\textbf{38.3}	\\
    6	&	multi-proj CTC + smoothing	&	33.2	&	33.1	&	39.1	\\
    7	&	Mutli gating + context	&	33.3	&	\textbf{32.9}	&	38.8	\\
    8	&	Mutli gating + smooth	&	33.5	&	33.5	&	39.2	\\
    \bottomrule
    \end{tabular}
    \label{tab:results3}
\end{table*}

Given there is almost a 19\% increase in the number of parameters, with no improvement, we realised it was important for parameters to be both language specific and shared parameters. So we re-implement gating to allow only a few layers to be gated, and other layers to be regular attention (see section \ref{sec:regular_attention}).

\begin{figure}[h!]
    \centering
    \includegraphics[scale=0.7]{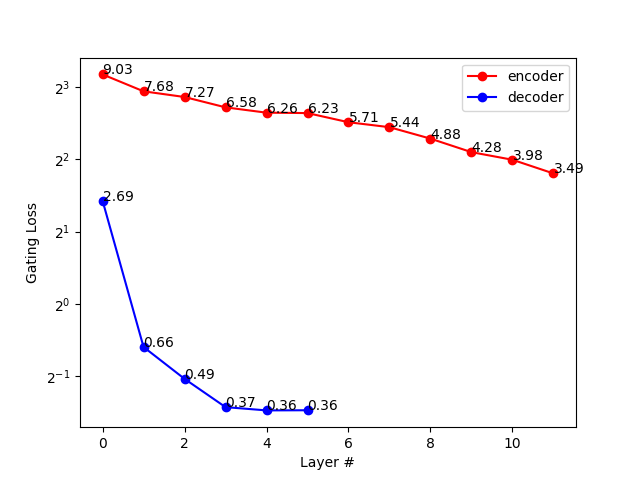}
    \caption{Average Gating Loss vs Layer \#}
    \label{fig:loss_graph}
\end{figure}

First, some terminology for clarity. Layers of the encoder that are closer to the input speech are referred to as bottom layers, and closer to the decoder as top layers. Similar convention for the decoder, data flow is from bottom to top. Our analysis of a trained model showed the absolute CTC gating loss shows a decreasing trend starting from the bottom to the top layers. This is consistent for layers in the encoder and the decoder (see figure \ref{fig:loss_graph})

This means that while later layers (that capture several hidden features of the encoded speech and text) are able to fit better to the language ID context, earlier layers act as conflicting parameters. Hence when applying gating only to a few layers, we focus on applying it to only the top few layers.

At this point of experiments we also focus on character-level gating. We take this step due to reasons mentioned in our qualitative analysis of gating weights (section \ref{sec:gating_weights}). But this brings us a new set of challenges that we discussed in section \ref{sec:stc_loss}. With word-level LID, the output sequences were small enough that could align using CTC loss. However character-level language ID data is roughly 5-6 longer than word-level, and CTC often results in \texttt{nan/inf} loss, especially in the decoder.

For ease of understanding the following abbreviations are used - \textbf{L2 enc/dec} is last 2 layers of gating in encoder (or decoder respectively, \textbf{F2} for first 2). All experiments from here on use character-based LID, unless mentioned otherwise. (table \ref{tab:results4})

\begin{table*}[t!]
    \caption{WER in \%: Method 2 - Partial gating, CTC trim}
    \centering
    \begin{tabular}{c l r r r}
    \toprule
    \textbf{S.no} & \textbf{Name} &	\textbf{d-cs} &	\textbf{t-cs} &	\textbf{d-mono}	\\
    \midrule
    1	&	Baseline	&	\textbf{32.6}	&	32.9	&	\textbf{38.3}	\\
    9	&	Word LID, L2 enc + L2 dec	&	32.8	&	\textbf{32.5}	&	38.5	\\
    10	&	Char LID, L2 enc + L2 dec	&	38	&	37.8	&	43	\\
    11	&	only encoder	&	34.8	&	34.5	&	40.1	\\
    12	&	encoder CTC trim + decoder CTC trim	& 33.1 & 33.3 &	38.9 \\
    13	&	encoder CTC trim + decoder CE	&	34.8	&	34.4	&	40	\\
    \bottomrule
    \end{tabular}
    \label{tab:results4}
\end{table*}

Since we did not see improvement in performance from Method 1, we move on to Method 2. We continue use partial gating on the last 2 layers, and introduce the STC loss into our network optimization. However we only perform a few experiments with it, as it takes longer to train. Wherever CTC loss is used, random trimming is also added to avoid inf loss issues. (table \ref{tab:results5})

\begin{table*}[t!]
    \caption{WER in \%: Method 2 - STC and gating loss weight}
    \centering
    \begin{tabular}{c l r r r}
    \toprule
    \textbf{S.no} & \textbf{Name} &	\textbf{d-cs} &	\textbf{t-cs} &	\textbf{d-mono}	\\
    \midrule
    1	&	Baseline	&	\textbf{32.6}	&	\textbf{32.9}	&	38.3	\\
    14 
    &	L2 enc CTC + L2 dec CE ($w_g = 0.2$)	&	32.9 & 33.4	& 38.9	\\
    15 
    &   L2 enc \textbf{STC} + L2 dec CE ($w_g = 0.2$)	&	33.3	&	33.6 &	39.4	\\
    16 
    &	L2 enc CTC + L2 dec \textbf{CTC} ($w_g = 0.2$) &	33.6	&	33.5 &	39.5	\\
    17 
    &	L2 enc CTC + L2 dec \textbf{STC} ($w_g = 0.2$)	&	33.3	&	33.5 &	39.0	\\
    18 
    &	L2 enc CTC + L2 dec CE ($w_g = 0.5$)	&	33.3	&	33.1 &	\textbf{38.1}	\\
    19 
    &	\textbf{L4} enc CTC + \textbf{L4} dec CE ($w_g = 0.5$)	&	33.2	&	33.0 &	38.7	\\
    20 
    &	L2 enc CTC + \textbf{F2 dec} CE ($w_g = 0.5$)& 33.4&	33.3	&	39.2	\\

    21	&	L2 enc CTC + L2 dec CE ($w_g = 0.75$)	&	34.1	&	34.1	&	39.8	\\
    \bottomrule
    \end{tabular}
    \label{tab:results5}
\end{table*}

The bad performance in experiment 10 can be explained by the inf loss (from CTC). The experiments after that focus on mitigating this issue.

As our qualitative analysis will show ahead (section \ref{sec:quali}), the model was successfully able to attend to specific parameters for their respective languages using the no-projection technique, but no improvement in WER. One suspect could be over-training and over optimization for the LID part of loss term. So we attempt to reduce the weight of the loss term as training progresses, in a step-wise manner. We'll represent these two experiments in the form of $w_g = x::y::z$ representing $w_g = x$ for 10 epochs, $w_g = y$ for another 10, and $w_g = z$ for the remaining time, etc. Both experiments use CTC (with trimming) for the encoder, and CE loss for the decoder (table \ref{tab:results6})

\begin{table*}[t!]
    \caption{WER in \%: Method 2 - reduction of gating weight periodically}
    \centering
    \begin{tabular}{c l r r r}
    \toprule
    \textbf{S.no} & \textbf{Name} &	\textbf{d-cs} &	\textbf{t-cs} &	\textbf{d-mono}	\\
    \midrule
    1	&	Baseline	&	\textbf{32.6}	&	32.9	&	\textbf{38.3}	\\
    22 & L2 enc CTC + L2 dec CE ($w_g = 1.::0.1::0.$) & 33.3 & 33.1 &	38.9 \\
    23	&	L2 enc CTC + L2 dec CE ($w_g = 1.::0.1$) & 32.9 & \textbf{32.8} & 38.7 \\
    \bottomrule
    \end{tabular}
    \label{tab:results6}
\end{table*}

Next we try unsupervised training of the gating weights, by setting $w_g$ to zero, and letting the parameters automatically guess the language of the speech, similar to \cite{lu2020bi}. However, qualitative analysis showed that the gating weights were random for this experiment, and it did not show better performance either (see table \ref{tab:results7}). This is different from Lu et al's paper as those gating weights had pre-trained expert encoders to rely on.

\begin{table*}[t!]
    \caption{WER in \%: Method 2 - Supervised vs Unsupervised training}
    \centering
    \begin{tabular}{c l r r r}
    \toprule
    \textbf{S.no} & \textbf{Name} &	\textbf{d-cs} &	\textbf{t-cs} &	\textbf{d-mono}	\\
    \midrule
    1	&	Baseline	&	\textbf{32.6}	&	\textbf{32.9}	&	\textbf{38.3}	\\
    24	&	L2 enc CTC + L2 dec CE ($w_g = 1$)	&	33.5	&	33.4	&	39.3	\\
    25	&	L2 enc CTC + L2 dec CE ($w_g = 0$)	&	33.1	&	33.5	&	38.9	\\
    \bottomrule
    \end{tabular}
    \label{tab:results7}
\end{table*}

So learning from \cite{lu2020bi} and our experiments, we try out pre-training these models using mixed monolingual data, and then finetuning with the 100 hours of code-switched data. The monolingual data includes the 100 hours of Gujarati data we were using before, along with 100 hours of English data. We also try training with all 300 hours of data together. (see table \ref{tab:results8} and \ref{tab:results9})

\begin{table*}[t!]
    \caption{WER in \%: Method 2 - pretraining with monolingual data}
    \centering
    \begin{tabular}{c l l r r r}
    \toprule
    \textbf{S.no} & \textbf{Name} & \textbf{Mode} &	\textbf{d-cs} &	\textbf{t-cs} &	\textbf{d-mono}	\\
    \midrule
    26	&	Baseline	& pretraining &	56.9	&	55.8	&	45.6	\\
    	&		& finetuning & \textbf{34.1	}&	\textbf{34.0}	&	43.2	\\
    (old)  &  \textit{Baseline } & \textit{(only CS)}& \textbf{34.1} & 34.2 & 46.6 \\
    27	&	L2 enc CTC + L2 dec CE & pretraining &	55.0	&	53.9	&	43.9	\\
    	&	& finetuning &	36.5	&	36.3	&	\textbf{42.4}	\\
    \bottomrule
    \end{tabular}
    \label{tab:results8}
\end{table*}

The finetuning baseline seems to have similar WER on code-switched data in comparison to experiments with only code-switched data. This indicates that in this setting, the monolingual data is not helping the model in code-switched setting at all. We do notice some improvement in monolingual (Gujarati) performance when using gating and language specific parameters. We also note that in both models, finetuning with code-switched speech only improves performance on monolingual examples.

We finally attempt to increase performance with this additional data by reducing the monolingual English data we give to the model. We retain only 27 hours of short samples of English speech, as this is much more representative of the total amount of English data in the code-switched dataset (English being the embedded language). $w_g = 0.5$ (see table \ref{tab:results9})

\begin{table*}[t!]
    \caption{WER in \%: Method 2 - Limited English Data}
    \centering
    \begin{tabular}{c l r r r}
    \toprule
    \textbf{S.no} & \textbf{Name} &	\textbf{d-cs} &	\textbf{t-cs} &	\textbf{d-mono}	\\
    \midrule
    1	&	Baseline \textit{(200 hours)}	&	\textbf{32.6}	&	32.9	&	\textbf{38.3}	\\
    28	&	Baseline \textit{(300 hours)}  &	33.2	& 33.0	& 39.0	\\
    29	&	Baseline \textit{(227 hours)} &	\textbf{32.6} & \textbf{32.8} & 39.0 \\
    30	&	L2 enc CTC + L2 dec CE \textit{(227 hours)} &	33.2 & 33.1 & 39.3	\\
    31 & L2 enc CTC + L2 dec CE \textit{(300 hours)} & 33.6 & 33.5 & 39.0 \\
    32	&	L6 enc CTC + L3 dec CE \textit{(227 hours)} &	34.0 &	34.0	&	39.6	\\
    \bottomrule
    \end{tabular}
    \label{tab:results9}
\end{table*}

\section{Other Evaluation Techniques}

CM-WER \cite{Taneja2019ExploitingMS} quantifies our task well. CM-WER first picks the indices of words that precede or follow a word of a different language. Then the error rate is calculated as the number of such words predicted incorrectly (substituted or deleted or inserted) vs total number of such words.

Firstly, we clarify how the metric to represent word error rate. We know the formula for WER is
$$\text{WER} = \frac{S + D + I}{S + D + C}$$

By the definition of CM-WER, we simply add the number of mistakes ($M = S + D + I$) at code switched points, add the number of correct such positions ($N = C$) and calculated it as
$$\text{CM-WER} = \frac{M}{M + N}$$

However, in case of code-switching points, we can have no insertions, as the code-switch point is explicitly marked by a reference word in a particular language, which cannot be empty. Hence with $I = 0$, both these metrics are the same.

Evaluations of several of our experiments show that there is \textit{worsening} of code-switch point performance in gating methods. We also saw an even worsening of non-code switch point error rate (i.e. words in the reference which are surrounded by other words of the same language). The numbers in table \ref{tab:results_cmwer} shows high error-rate for non code-switch points as all insertion errors from the model are included in this. The \textbf{S.no} can be matched to earlier experiments, roughly sorted by type.

\begin{table*}[t!]
    \caption{WER (in \%) at Code-Switch Points and non Code-Switch Points}
    \centering
    \begin{tabular}{c l c c c c}
    \toprule
    \textbf{S.no} & \textbf{Name} &	\textbf{d CM} & \textbf{d non-CM} & \textbf{t CM} &	\textbf{t non-CM}	\\
    \midrule
    1 & Baseline	& 27.17 & \textbf{36.45} & \textbf{27.88} & 36.32	\\
    28 & Baseline \textit{(300 hours)}&	27.70	&	37.08	&	27.99	&	36.40	\\
    29 & Baseline \textit{(227 hours)}	&	\textbf{26.65}	&	36.86	&	\textbf{27.88}	&	\textbf{36.18}	\\
    \midrule
    & \textit{Method 1 \& Word LID gating} \\
    \midrule
    2 & Shared gating	&	28.88	&	36.95	&	28.81	&	36.57	\\
    3 & Multi gating	&	28.01	&	\textbf{36.24}	&	28.00	&	36.15	\\
    4 & Gating with multi-proj CTC	&	28.38	&	37.43	&	28.91	&	36.88	\\
    9 & L2 e-CTC + L2 d-CTC	&	\textbf{27.88}	&	36.28	&	\textbf{27.93}	&	\textbf{35.61}	\\
    \midrule
    & \textit{Method 2 \& Char LID gating (CTC with trim)} \\
    \midrule
    18 & L2 e-CTC + L2 d-CE ($w_g = 0.5$)	&	28.02	&	36.98	&	28.54	&	36.18	\\
    30 & L2 e-CTC + L2 d-CE ($w_g = 0.5$, \textit{227 hours})	&	28.39	& 36.65	& 28.75 &	36.12 \\
    31 & L2 e-CTC + L2 d-CE ($w_g = 0.5$, \textit{300 hours})&	27.99	&	37.53	&	28.74	&	36.79	\\
    24 & L2 e-CTC + L2 d-CE ($w_g = 1$)	&	27.79	&	37.47	&	28.63	&	36.58 \\
    25 & L2 e-CTC + L2 d-CE ($w_g = 0$) & 28.25 & \textbf{36.57} & 28.48 & 36.99 \\
    14 & L2 e-CTC + L2 d-CE ($w_g = 0.2)$	&	28.09	&	36.23	&	28.81	&	36.57	\\
    16 & L2 e-CTC + L2 d-\textbf{CTC} ($w_g = 0.2$, trimmed) &	28.45	&	37.31	&	29.54	&	36.17	\\
    17 & L2 e-CTC + L2 d-\textbf{STC} ($w_g = 0.2$)	&	28.15	&	36.97	&	28.75	&	36.76	\\
    20 & L2 e-CTC + F2 d-CE ($w_g = 0.5$) &	28.14	&	37.11	&	28.76	&	36.48	\\
    15 & L2 e-\textbf{STC} + L2 d-CE ($w_g = 0.2$)	&	28.28	&	36.92	&	28.63	&	37.04	\\
    22 & L2 e-CTC + L2 d-CE ($w_g = 1.::0.1$) &	\textbf{27.64}	&	36.66	&	\textbf{28.21}	&	\textbf{36.00}	\\
    23 & L2 e-CTC + L2 d-CE ($w_g = 1.::0.1::0.$) &	28.19	&	36.88	&	28.75	&	36.07	\\
    19 & \textbf{L4} e-CTC + \textbf{L4} d-CE ($w_g = 0.5$)&	28.12	&	36.76	&	28.57	&	36.10	\\
    32 & \textbf{L6} e-CTC + \textbf{L3} d-CE (\textit{227 hours}) &	29.24	&	37.44	&	29.64	&	37.04	\\
    \midrule
    & \textit{Method 2 - Pretraining + finetuning} \\
    \midrule
    26.a & Baseline	(\textit{200 hours monolingual})&	70.12	&	47.64	&	70.12	&	46.30	\\
    26.b & Baseline (\textit{finetuned on 100 hours CS})	&	\textbf{27.96}	&	\textbf{38.42}	&	\textbf{28.47}	&	\textbf{37.77}	\\
    27.a & L2 e-CTC + L2 d-CE ($w_g = 0.5$, \textit{200 hours})	&	69.26	&	44.93	&	69.35	&	43.39	\\
    27.b & L2 e-CTC + L2 d-CE ($w_g = 0.5$, \textit{100 hours CS})	&	32.32	&	39.53	&	32.62	&	38.84	\\
    \bottomrule
    \end{tabular}
    \label{tab:results_cmwer}
\end{table*}

\section{Qualitative Analysis}
\label{sec:quali}

We see no significant improvement in WER using our methods. We suspect that one reason for this is a noisy dataset. We listened to some samples from the training data, and despite not knowing Gujarati, we all agreed that the speech was very fast and at times incomprehensible. Some reference texts were incomplete, and our model(s) often was able to fill in the remaining dialogue. Inspite of this or as a result of this, we notice a few good and bad outcomes from our model.

\subsection{Analysis of gating weights}
\label{sec:gating_weights}

In the word-level LID experiments - we noticed that the probability distribution had a huge bias towards predicting $\langle\text{BLANK}\rangle (\epsilon)$ tokens, and learns to predict peaks for each word in the output sequence. This led to likely output sequence like "$\epsilon\epsilon\langle\text{G}\rangle\epsilon\epsilon\langle\text{E}\rangle\epsilon\epsilon\epsilon\langle\text{G}\rangle$" for a 3 word prediction. This was not what was intended, and we attributed this to two factors - smaller output sequences, and how CTC loss is designed. Which is why we moved to longer character-level LID, and STC loss.

Here we notice that the gating weights from each gating layer (encoder) is very well aligned to a rough interpolation of the LID. We do not show plots for decoder as it is a beam search based model, but our analysis of a few example showed similar knowledge of the language. This is significant because these gating weights are calculated per frame from just the speech encoding.

\begin{figure}[h!]
    \centering
    \includegraphics[scale=0.5]{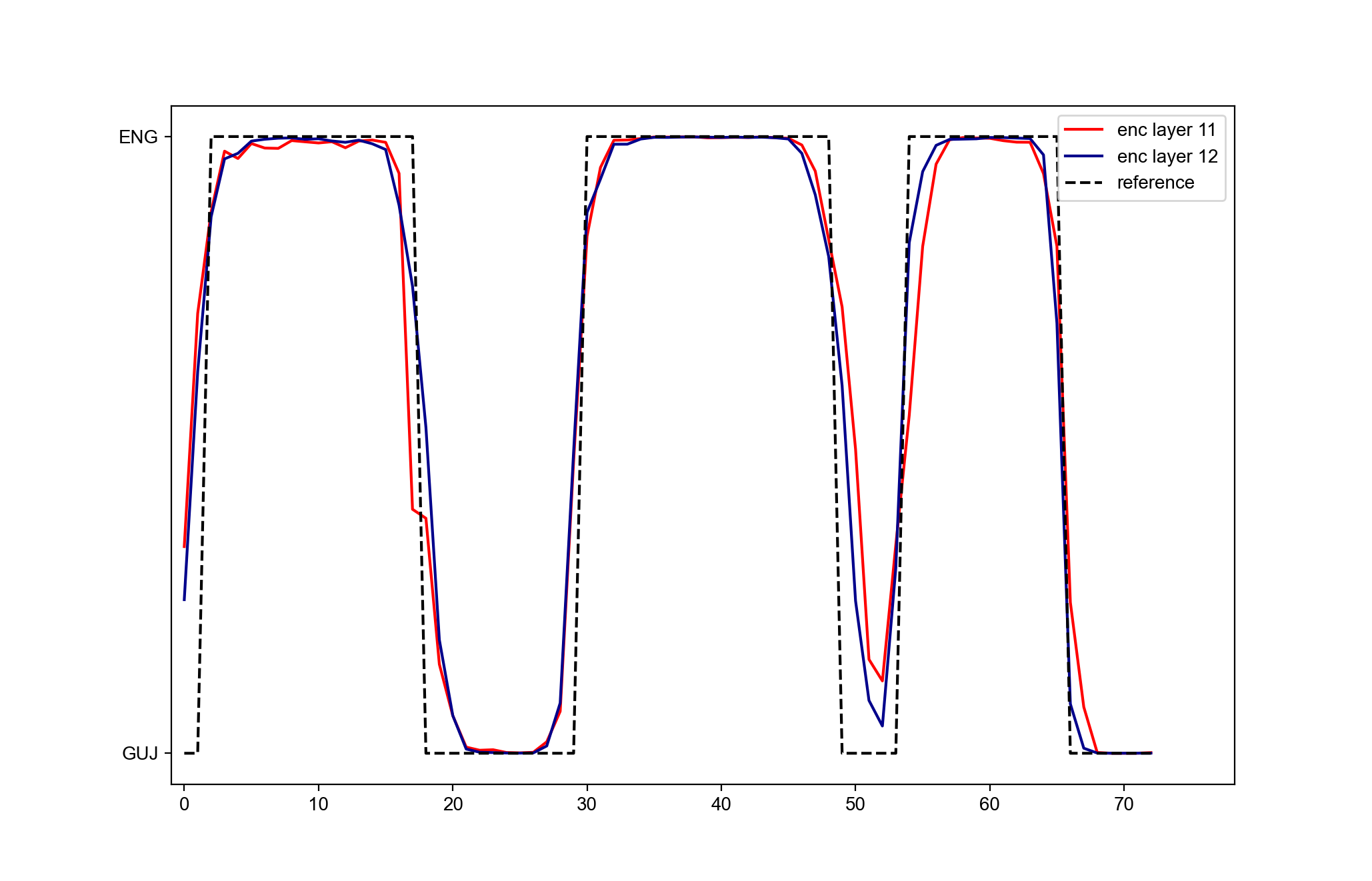}
    \caption{Gating weights alignment to reference LID}
    \label{fig:align}
\end{figure}

The text for the example in figure \ref{fig:align} is:
\begin{figure*}[h!]
    \includegraphics[scale=0.5]{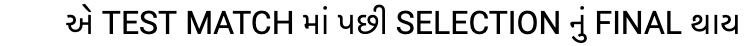}
    \label{fig:my_label}
\end{figure*}
\vspace{-10pt}

which roughly translates to "The final selection will be after this test match".

\subsection{Similar output from language specific parameters}

Despite the model learning how to predict the language of speech, it was predicting the same text for either set of language specific parameters. We run this experiment by setting the gating weights to all 1 and 0, then vice versa, and see no noticeable change in the decoding logs. We conclude that both set of language parameters are very close to each other, therefore model is unable to use the LID prediction. Further to this, since each set of parameters only see gradients from parts of the encoding, either of them are not fully trained with all data, and hence often perform poorly.

%% file: chap_06_efficiency.tex
\chapter{\textsc{Efficiency Improvements in STC Loss}}
\label{ch:effi}

The following is a naive algorithm for calculating STC loss. We split the loss formula mentioned in the previous chapter into the first repeating prefix in the output sequence $\mathbf{y}$ and the rest. We then get recursive loss computations, transform them back to probability by negative exponentiation, take a sum over all prefixes of input $\mathbf{a}$ that could fit, and take the negative log to get the loss.

\section{Algorithm}
\lstset{language=Python, morekeywords={STCLoss}}
\lstset{frame=lines}
\lstset{caption={Naive algorithm for STC Loss}}
\lstset{label={lst:code_direct}}
\lstset{basicstyle=\footnotesize\ttfamily,breaklines=true}
\begin{lstlisting}
def STCLoss(W, Y):
    assert len(W) >= len(Y)
    if len(Y) == 0:
        return inf if len(W) != 0 else 0
    prefixLength, prefix = NumSamePrefix(Y)
    lenRemaining = len(Y) - len(W)
    losses = []
    for x in range(len_remaining + 1):
        negLogProb = sum(W[:prefixLength + x] == prefix)
        remLoss = STCLoss(W[prefixLength + x:], Y[prefixLength:])
        normLoss = log(prefixLength)
        losses.append(negLogProb + remLoss + normLoss)
    return -LogSumpExp(-losses)

\end{lstlisting}

here W is the negative log probabilities of each character in the input alignment (-log(a)) and Y is the output character sequence. \texttt{logSumExp} is as the name suggests, and \texttt{numSamePrefix} returns the next character in the output sequence and how many times it repeats ($\geq 1$). \texttt{W[i] == char} is the negative log probability of character `char' occuring at position `i'.

There are two ways to implement this dynamically. Top down, and bottom up.

Top down or memoization is starting from the original task ($\text{loss}(x, y)$) and calling the sub-problems utmost once, storing the values for repeated calls. This is at times inefficient depending on the implementation language, but only computes the required sub-problems.

Bottom up or tabulation is enumerating and storing all subsequences of input that can align with output. This may have extra sub-problem computations that were not required by the task, but subsequently is heavily parallelizable on the platform we're concerned with - pyTorch.

Find additional details of optimization of this code in Appendix \ref{sec:opti}.

%% file: chap_07_Summary_and_Conclusions.tex
\chapter{\textsc{Summary and Conclusion}}

Our task was to improve code-switching speech recognition on Gujarati-English speech with roughly 200 hours of parallel data. We analysed new avenues of improving accuracy and found that transliterating the model's output and reference lead to small but significant decrease in error. We experiment with explainable ideas that help the model learn the spoken language, and not just the phonetic characteristics from the speech. However, none of our experiments manage to reduce error rate significantly.

\section{Contributions}
During the course of this work, we contribute the following methods and analysis to the ASR community, despite the lack of results yet

\begin{enumerate}
    \item An early iteration of introducing language dependencies within the encoder (and decoder) of an end-to-end model, and push towards explainability of these models.
    \item A new soft temporal alignment loss: \textbf{Seamless Temporal Classification}, inspired by CTC Loss.
    \item We made certain implementation and formula based changes in Code-Mixed Word Error Rate.
\end{enumerate}

We also add a detailed explanation of the pipeline of ESPnet (version 2) (in Appendix \ref{ch:pipeline}). It is discussed in the context of loading data into the pipeline, and the modification needed for it.

\section{Scope for future research}
\begin{itemize}
    \item STC loss could be used for direct ASR optimization. This could include a different ``prior" over the projection $\B(a)$ rather than the uniform distribution used. Or just output the smallest projection, as a lot of languages (e.g. Hindi, Gujarati or Korean) rarely have repeating consecutive characters, and hence separate modelling for them need not be necessary for good predictions of spoken text. For e.g. consecutive repeating characters make up 2.8\% of a lexicon, while it makes up only 0.4\% of a Gujarati lexicon. Even if significant, it can be fixed by a reconstruction model \cite{diwan2020reduce}.
    \item Experiments with more than 2 languages might show better performance.
    \item This kind of setup with the right tuning could be used for speech translation/transliteration. However this is also achievable with traditional transliteration modules cascaded on an ASR model.
    \item Instead of interpolating features in method 2 (see section \ref{sec:gating_method2}) we could also sample from the probability distribution and select output accordingly, dropping the other language per frame entirely.
\end{itemize}

%% file: Appendix_I.tex
\chapter{\textsc{Pipeline of ESPnet v2.0}}
\label{ch:pipeline}
\section{How to load additional data}

The following is a detailed explanation of how to load additional data apart from speech and text, in this case language ID text, in one of the commonly used neural network based speech recognition toolkit - ESPnet \cite{watanabe2018espnet}. Specifically, we used version 2 of the library. While it does derive a lot from kaldi\cite{povey2011kaldi}'s style of data format, the difference in version 2 is that it doesn't depend on kaldi's binaries for it's pre-processing.

Before you read on further, keep in mind that this is simply section discusses loading raw data using the existing loaders. Much of the process was simplified as the task of loading lang\_id is similar to the text file loaded (kaldi style format)

\texttt{train\_one\_epoch} in \texttt{espnet2/train/trainer.py} is our center of attention. it is here in a loop that batches are obtained from an \texttt{iterator}. The \texttt{reporter} is simply a wrapper.
This function is called from run, iterator is passed to it, which is built from \linebreak \texttt{train\_iterator\_factory.build\_iter}, which is passed the epoch \# to set random seed.

\texttt{train\_iterator\_factory} is an object returned by \texttt{build\_multiple\_iter\_factory}, of \texttt{cls} (of type ASRTask). It's of type MultipleIterFactory, which takes \texttt{build\_funcs} as an initialization argument.

\texttt{build\_funcs} is a list of partial functions (similar to lambda functions, each is fully given all arguments relating to speech file path, text file path, and in this case, additionally, language id file). The partial is used to maintain laziness. The number of build\_functions in build\_funcs is the number of splits made in data processing.

Each \texttt{build\_func} when called returns a iterator factory, built using a subset of data - a bunch of speech, text, etc. examples. This \texttt{build\_iter} returns a \texttt{torch.DataLoader}.

The \texttt{collate\_fn} to this DataLoader converts a list of dictionaries, each having single speech, text, (and language id) tensors to a dictionary of batched tensors of speech, text (and language id). The speech is pre-processed through the frontend in Stage 9, and hence is not trained along with the rest of the ASR model. The magic of data processing (on the fly) hence happens in the \texttt{ESPnetDataset} class. The \texttt{ESPnetDataset} is constructed by taking the path list (as described above) with their keys for the dictionary that will be eventually made  (e.g. [(`wav.scp', `speech', `sound')]). Here `sound' is the data type and `speech' is the key.

A side note: post collate - the dictionary that is created is sent as keyword arguments to the ESPnetModel's forward. So a modification in what data is being sent would require a modification in the template of the forward function. The default in their code is ``speech" and ``text", which is accompanied by ``speech\_lengths" and ``text\_lengths"

Accompanying the data info is a preprocess function, which is where the conversion of string to numbers for the model happens.
So to add data, one would modify the preprocess function in order to accommodate our new data type, and modify the asr.sh script to add this additional data in the format shown in the examples.


%% file: Appendix_II.tex
\chapter{\textsc{Incremental Optimization of Code}}
\label{sec:opti}
We compare the average time taken for calculating the CTC loss and the STC loss on a batch size of 5, with a random input and repetitive language ID style output. It takes 1.3 seconds for CTC to perform a forward and backward 50 times (50 iterations) on this input, with one simple FF layer.

We initially implemented the process top down, with \texttt{KeyError} in python's default dictionary helping us solve first-time computations. This proved to be expensive. One forward loss takes roughly 30 seconds time for the same test input.

We then moved to bottom up, tabulating results for all O(NM) sub-problems in this task. This reduced the time taken to 3 seconds per forward pass.

This code however had 3 loops, one on the output sequence and one on the input sequence (for each sub-problem), plus another on the input sequence within the sub-problem itself. By parallelizing one of the input loops, we bring down the time required for 50 iterations to 51 seconds. Reducing another loop on the input sequence, this is brought down to 7 seconds. The loop on the output sequence cannot be removed due to data dependency.

This was still severely inefficient compared to CTC, and on further inspection using cProfile, the bottleneck was a very inefficient implementation of \texttt{torch.flip}. Our use case for it was a reversed version of \texttt{torch.cumsum}, which is also highlighted in \href{https://github.com/pytorch/pytorch/issues/33520}{Github - pyTorch issue \# 333520}. As in the solution mentioned, we get rid of the use of \texttt{torch.flip}, and with a few other tweaks, we managed to get the computation time down to 3 seconds.

It's difficult to further reduce this, as we're comparing performance to CTC written in C++, pre-compiled and optimized. STC loss also fairs much poorly when the number of repetitions are reduced in the output sequence.
Despite all the efficiency improvements, training time with STC loss is almost double in comparison to CTC Loss, and it's the bottleneck in the training routine.